\documentclass[runningheads]{llncs}

\usepackage{graphicx}
\usepackage{varioref}
\usepackage{url}
\usepackage{epsfig}
\usepackage{amsmath}
\usepackage{epsfig}
\usepackage{amssymb}
\usepackage{graphicx}
\usepackage{algorithm2e}
\usepackage{algorithmic}
\usepackage{subfigure}
\usepackage{float}
\restylefloat{table}
\def\urltilda{\kern -.15em\lower .7ex\hbox{\~{}}\kern .04em}

\newtheorem{thm}{Theorem}
\newtheorem{ex}{Example}
\newcounter{asm}
\newenvironment{asm}{%
  \par\medskip\refstepcounter{asm}%
  \noindent\textbf{Assumption \theasm:}\quad}{\par\medskip}
\labelformat{asm}{\theoremname~#1}
\newcommand\numberthis{\addtocounter{equation}{1}\tag{\theequation}}

\begin{document}

\pagestyle{headings}


\title{Consensus-Based Modelling using Distributed Feature Construction\thanks{A short 6-page version of this paper was presented at the 24th International Conference on Inductive Logic Programming, held in conjunction with ECML PKDD, France. Furthermore, a significant part of the work in this paper was done when the first author was an Associate Research Scientist at the Center for Computational Learning Systems (CCLS), Columbia University, NY.}}

\titlerunning{Consensus-Based Modelling}        

\author{Haimonti Dutta\inst{1} \and Ashwin Srinivasan\inst{2}}


\institute{Department of Management Science and Systems  \\
            University at Buffalo \\
            New York, NY 14260. \\
            \email{haimonti@buffalo.edu} \\
           \and
           Department of Computer Science \\
           IIIT, Delhi.\\
           \email{ashwin@iiitd.ac.in}
}


\maketitle

\begin{abstract}
A particularly successful role for Inductive Logic Programming (ILP)
is as a tool for discovering useful relational features for subsequent use in a
predictive model. Conceptually, the case for using ILP to construct relational
features rests on treating these features as functions, the automated discovery of which
necessarily requires some form of first-order learning. Practically,
there are now several reports in the literature that suggest that
augmenting any existing features with ILP-discovered relational features can
substantially improve the predictive power of a model.
While the approach is straightforward enough,
much still needs to be done to scale it up to explore more fully the space
of possible features that can be constructed by an ILP system. This
is in principle, infinite and in practice, extremely large. Applications have been
confined to heuristic or random selections from this space. In this paper,
we address this computational difficulty by allowing features to
be constructed in a distributed manner. That is, there is a network of
computational units, each of which employs an ILP engine to construct
some small number of features and then builds a (local) model. We then employ a consensus-based algorithm, in which
neighbouring nodes share information to update local models. For a category of models
(those with convex loss functions), it can be shown that the algorithm will result in all nodes
converging to a consensus model. In practice, it may be slow to achieve this convergence.
Nevertheless, our results on synthetic and real
datasets that suggests that in relatively short time
the ``best'' node in the network
reaches a model whose predictive accuracy is comparable to
that obtained using more computational effort in a non-distributed setting
(the best node is identified as the one whose weights converge first).

\end{abstract}

\section{Introduction}
\label{intro}

The field of Inductive Logic Programming (ILP) has made steady
progress over the past two decades, in advancing the theory,
implementation and application of logic-based relational learning.
A characteristic of this form of machine-learning is that
data, prior knowledge and hypotheses are usually---but not always---expressed in a
subset of first-order logic, namely logic programs.
Side-stepping for the moment the question ``why logic programs?'', it is evident
that settling on some variant of first-order logic allows the construction of tools that
enable the automatic construction of descriptions that use relations
(used here in the formal sense of a truth value assignment to $n$-tuples).

There is at least one kind of                                          
tasks where some form of relational learning would appear to be necessary.
This is to do with the identification of functions
(again used formally, in the sense of being a uniquely
defined relation) whose domain is the set of instances in the data.
An example is the construction of new ``features'' for data analysis
based on existing relations (``$f(m) = y$ if a molecule $m$ has 3 or more
benzene rings fused together otherwise $f(m) = n$'').
Such features are not intended to constitute a stand-alone
description of a system's structure. Instead, their purpose is to enable
different kinds of data analysis to be performed better. These may be constructing models
for discrimination, joint probability distributions,
forecasting, clustering, and so on. If a logic-based relational learner like an ILP
engine is used to construct these relational features,
then each feature is formulated as a logical formula.
A measure of comprehensibility will be retained in the resulting models that use
these features.

%
%
\noindent
%
%

The approach usually, but not always, separates relational learning
(to discover features) and modelling
(to build models using these features). There will of course
be problems that require the joint identification of relational
 with features and models---the emerging area of statistical relational
learning (SRL), for example, deals with the conceptual and implementation issues
that arise in the joint estimation of statistical parameters and relational
models. It would appear that separate construction of features
and statistical models would represent no more than a poor man's SRL.
Nevertheless, there is now a growing body of
research that suggests that augmenting any existing
features with ILP-constructed relational ones can substantially
improve the predictive power of a statistical model
(see, for example: \cite{JoshiRS08,Amrita12,Specia_09,RamakrishnanJBS07,SpeciaSRN06}).
There are thus  very good practical reasons to persist with this variant of
statistical and logical learning for data analysis.

There are known shortcomings with the approach which can limit its
applicability. First, the set possible relational features
is usually not finite This has led to an emphasis on syntactic and
semantic restrictions constraining the features to some finite set. In practice
this set is still very large, and it is intractable to identify an optimal
subset of features. ILP engines for feature-construction therefore employ some form
of heuristic search.
Second, much needs to be done to scale ILP-based feature discovery
up to meet modern ``big'' data requirements. This includes the abilities to
discover features using very large datasets not all stored in one place,
and perhaps only in secondary memory; from relational data arriving in a streaming manner;
and from data which do not conform to expected patterns (the concept changes, or
the background knowledge becomes inappropriate).
Third, even with ``small'' data, it is well-known that obtaining the
value of a feature function for a data instance can be computationally hard.
This means that obtaining the feature-vector representation using ILP-discovered
features can take large amounts of time.

\begin{figure}[htb]
\centerline{\includegraphics[height=0.30\textheight]{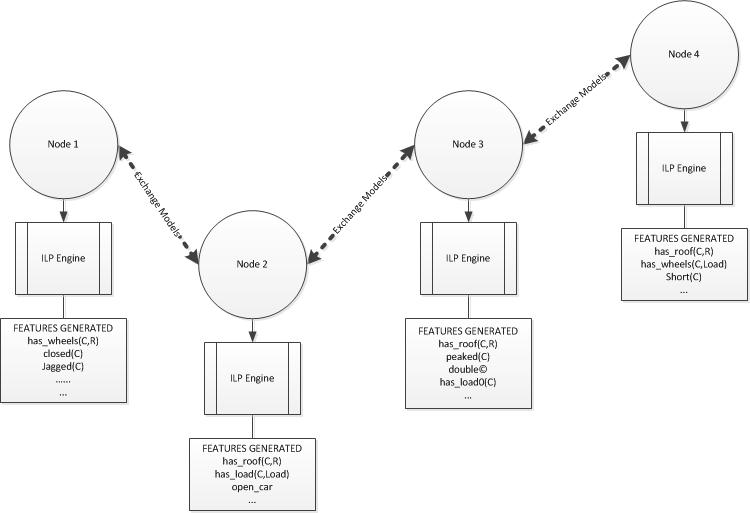}}
\caption{An illustrative example of the consensus-based approach using Michalski's ``Trains" problem \protect{\cite{Larson_77}}.
Each node has local features generated from an ILP engine, builds local models, estimates loss and shares
information with its neighbours. Eventually we would like all nodes converge to the same model.}
\label{illEx}
\end{figure}

This paper is concerned only with the first of these problems, namely how to construct
models when feature-spaces are very large.
Each node has access to some (but not all)
relational features constructed by an ILP
engine, and constructs a local linear model.
Using a simple consensus-based algorithm, all nodes in the network converge on
the optimal weights for all the features. The setting is naturally amenable to
distributed learning, providing us with a mechanism of scaling-up the construction
of models using ILP-based feature discovery. Our approach does not result an optimal answer
to the feature-selection problem. However it does have some positive aspects: it provides one
way of distributing the computational task of feature-construction; and given multiple, possibly overlapping sets of
relational features, it provides one way of identifying the best model (for a specific category of models).
Figure~\ref{illEx} illustrates this. The algorithm for distributed feature estimation, described in
Section~\ref{sec:probdesc} is an implementation of such an approach.
%

%

Section \ref{sec:related} presents related work; 
Section \ref{sec:probdesc} formulates the problem as one of consensus-based model construction. Section \ref{sec:alg}
presents an iterative procedure for constructing models in a network of nodes capable
of exchanging information about their local models. Experimental
results are in Section \ref{sec:expt}. Section \ref{sec:concl} discussed open
issues and concludes the paper.

\section{Related Work}
\label{sec:related}

Techniques for selecting
from a (large) but finite set of features of known size $d$
has been well-studied within the machine learning, usually under
the umbrella-terms of filter-based or wrapper-based methods
(see for example, \cite{John_94,Liu_98}). 
While most of the early work was intended for implementation on a single
machine, feature-selection from fixed-sized feature spaces has been extended
to a distributed setting for large datasets. Here, the data are partitioned and
placed on different processors; processors must communicate to find parameters that
minimize loss over the entire dataset. However, communication has to be fast
enough so that network latencies do not offset computational gains. The partitioning
scheme -- horizontal (all instances have the same features) or
vertical (instances have access to only a subset of features) plays an
important role in the design of these distributed algorithms.
Early work in decentralized optimization was marked by interest in
consensus-based learning, distributed optimization and minimization with the
seminal work of Bertsekas, Tsitsiklis and colleagues
(\cite{Tsitsiklis_86,Tsitsiklis_84,Bertsekas_97}).
More recently, researchers have shown that convergence properties of these
decentralized algorithms can be related to the network topology by
using spectral properties of random walks or path averaging arguments
on the underlying graph structure
(\cite{Boyd_06,Shah_09,Dimakis_06,Benezit_10}).
Learning feature subsets in distributed environments using decentralized optimization
has become an active area of research
(\cite{Duchi_12,Agarwal_14a,Chris_08}) in recent years. 

Agarwal et al. \cite{Agarwal_14a} present a system and a set of techniques for
learning linear predictors with convex losses on terabyte sized datasets.
Their goal is to learn problems of the form
$ \text{min }_{w \in R^d} \sum_{i=1}^{n} l (w^T x_i; y_i) + \lambda R(w)$
where $x_i$ is the feature vector of the $i^{th}$ example, $w$ is the weight vector
and $R$ is a regularizer. The data are split horizontally and examples are
partitioned on different nodes of a cluster. Duchi et al.
\cite{Duchi_12} present a dual averaging sub-gradient method which
maintains and forms weighted averages of sub-gradients in the network. An
interesting contribution of this work is the association of convergence of the
algorithm with the underlying spectral properties of the network.
Similar techniques for learning linear predictors have been presented
elsewhere (\cite{Mangasarian_95,McDonald_10,zinkevich_10},
\cite{Hogwild_11,Boyd_11}). The algorithm presented in this paper
differs from this body of literature in that the data are split
\emph{vertically} amongst nodes in the cluster thereby necessitating a
different algorithm design strategy. In addition, this is a batch algorithm and
hence quite different from distributed online learning
counterparts (\cite{Dekel_12a,Langford_09a,bottou_2011}).

Das et al. \cite{Das_10}
show that three popular feature selection criteria -- misclassification gain,
gini index and entropy can be learnt in a large peer-to-peer network.
This is then combined with protocols for asynchronous distributed
averaging and the secure sum protocols to present a privacy preserving
asynchronous feature selection algorithm.



Existing literature on discovering a subset of interesting features from
large, complex search spaces such as those by ILP engines
adapt one of the following strategies:

\begin{enumerate}
 \item Optimally~\cite{l1Reg,cvpr2007,nips2004} or
    heuristically~\cite{JoshiRS08,Amrita12,NageshRCKDB12,ChalamallaNSR08,Specia_09,RamakrishnanJBS07,SpeciaSRN06}
    solve a discrete optimization problem.

\item Optimally~\cite{JawanpuriaNR11,NairSRK12} solve a convex
    optimization problem with sparsity inducing regularizers; 
 \item Compute all relational features that satisfy some quality criterion by
    systematically and efficiently exploring a prescribed
    search space~\cite{pei2004,subseqGap2006,bitSpade,antunes2003,han2005,rakesh1995,han2004,gehrke2002,rastogi1999}.
\end{enumerate}

\noindent
Again, much of this has been of a non-distributed nature, and
usually assume a bound on the size of the feature-space. The latter
is not the case for a technique like the one proposed in
\cite{JoshiRS08}. This describes a randomized local search based
technique which repeatedly constructs features
and then performs a greedy local search starting from this
subset. Since enumeration of all local moves can be prohibitively large,
the selection of moves is guided by errors made by the model constructed
using the current set of features. Nothing is assumed about the size
of the feature-space, making it a form of vertical partitioning of the kind
we are interested in.
Multiple random searches can clearly be
conducted in parallel (although this is not done in the paper). As with
most randomised techniques of this kind, not much can be said about the
final model.
    
Perhaps of most interest to the work here is the Sparse
Network Of Winnow classifers described in \cite{Roth_98,CCRR99}.
As it stands, this horizontally partitions the data into subsets, constructs
multiple linear models using Winnow's multiplicative update process, and
finally uses a majority vote to arrive at a consensus classification. This
would appear, on the surface to be quite different to what we propose here.
Nevertheless, there are reasons to believe that this approach can be usefully
extended to the setting we propose. It has been shown elsewhere that the
Winnow-based approach can be extended to an infinite-attribute setting \cite{Blum:winnow}. The
work in this paper shows that consensus linear models are possible when
convex cost functions are used. Finally, from the ILP-viewpoint, \cite{ashbain:stream} shows how
it is possible to construct Winnow-based models in an infinite-attribute
setting using an ILP engine with a stream-based model of the data. Taken together,
this suggests that a combination of the techniques we propose, and those in
\cite{CCRR99} can be used to develop linear models that can handle both
horizontal partitioning of the data and vertical partitioning of the feature-space.

\section{Consensus-Based Model Construction}

\subsection{Problem Description}
\label{sec:probdesc}
Let $M$ denote an $n \times m$ matrix with real-valued entries.  This matrix 
represents a dataset of $n$ tuples of the form $X_i \in \mathbb{R}^m, 1 \le i \le n$.    
Assume, without loss of generality, this dataset has been vertically distributed over
$k$ sites $S_1, S_2, \cdots, S_k$ i.e. site $S_1$ has $m_1 $ features, $S_2$ has $m_2$ features
and so on, such that $|m_1| + |m_2|+ \cdots + |m_k| = |m|$,
where $|m_i|$ represents the number of features at site $S_i$\footnote{In the more general setting,
Site $S_i$ has a random subset of features $m_i \subset m$.}. 
Let $M_1$ denote the $n \times m_1$ matrix representing the dataset held by $S_1$,
$M_2$ denote the $n \times m_2$ matrix representing the dataset held by $S_2$ and so on.
Thus, $M = M_1:M_2: \cdots : M_k$ denotes the concatenation of the local datasets. 

We want to learn a linear discriminative function over the data set $M$. The global function to be
estimated is represented by $f_{g} = M W_{g}^{T} $ where $W_g$ is assumed to be a $1 \times m$ weight vector.
If only the local data is used, at site $S_1$, the local function estimated would be $f_1 = M_1 W_{1}^{T}$.
At site $S_2$, the local function estimated would be $f_2 = M_2 W_{2}^{T}$.
The goal is to describe a de-centralized algorithm for computing the weight vectors at sites $S_1, \cdots S_k$
such that on termination $W_{1} \approx W_g[1:m_1], W_2 \approx W_g[1:m_2], \cdots W_k \approx W_g[1:m_k]$
where $W_g[1:m_i]$ represents the part of the global weight vector for the attributes stored at that site $S_i$.
Clearly, if all the datasets are transferred to a central location, the global weight vector can be estimated.
Our objective is to learn the function in the decentralized setting assuming that transfer of actual data tuples is
expensive and may not be allowed (say for example due to privacy concerns). The weights obtained at each site on
termination of the algorithm will be used for ranking the features.

\begin{algorithm}[t]
\small
{
\SetKwData{Left}{left}\SetKwData{This}{this}\SetKwData{Up}{up}
\SetKwFunction{Union}{Union}\SetKwFunction{FindCompress}{FindCompress}
\SetKwInOut{Input}{input}\SetKwInOut{Output}{output}

\KwIn{$n \times m_i$ matrix at each site $S_i$, $G(V, E)$ which encapsulates the underlying communication framework, $T: $ no of iterations }
\KwOut{Each site $S_i$ has $W_i \approx W_g[1: m_i]$ }
\BlankLine

 %
\For{t = 1 to T}
 {
  (a) Site $S_i$ computes $M_i W_i^T$ locally and estimates the loss function\;
  (b) Site $S_i$ gossips with its neighbors $S_j \in \{N_i\}$ and obtains $M_j W_j^T$ for each neighbor\;
  (c) Site $S_i$ locally updates its function estimate as $J_i^t = \alpha_{ii}(M_i W_i^T) + \alpha_{ji} (M_j W_j^T)$ \;
  (d) Update the local weight vectors using stochastic gradient descent as follows:
        $\frac{\partial L_p}{\partial W_i} = -X_p (Y_p - J_i^t(W_i^t(p)))$\;
  (e) If there is no significant change in the local weight vectors of one of the sites then stop
 }
 
\caption{Consensus-Based Modelling}
\label{alg:fs}
}
\end{algorithm}

\subsection{Algorithm}
\label{sec:alg}

 We state the following assumptions under which the distributed algorithm operates:

\begin{asm}{Model of Distributed Computation.}
The distributed algorithm evolves over discrete time with
respect to a ``global'' clock\footnote{Existence of this clock is of interest only for theoretical analysis.}.
Each site has access to a local clock. Furthermore, each site has its own memory and can perform local
computation (such as computing the gradient on its local attributes). It stores $f_i$, which is the estimated local
function. Besides its own computation, sites may receive messages from their neighbors which will help in
evaluation of the next estimate for the local function.
\end{asm}

\begin{asm}{Communication Protocols.}
Sites $S_i$ are connected to one another via an underlying communication framework
represented by a graph $G (V, E)$, such that each site
$S_i \in \{S_1, S_2, \cdots , S_k\}$ is a vertex and an edge $e_{ij} \in E$ connects sites $S_i$ and $S_j$.
Communication delays on the edges in the graph are assumed to be zero.
It must be noted that the communication framework is usually expected to be application dependent.
In cases where no intuitive framework exists, it may be possible to simply rely on the physical connectivity of the machines,
for example, if the sites $S_i$ are part of a large cluster.
\end{asm}

Algorithm~\ref{alg:fs} describes how the weights for attributes will be estimated using a
consensus-based protocol. There are two main sub-parts of the algorithm:
(1) Exchange of local function estimate and (2) Local update based on stochastic gradient descent. Each of these sub-parts are discussed in further detail below.   Furthermore, assume that $J: R^m \rightarrow [0, \infty]$ is a continuously differentiable nonnegative cost function with a Lipschitz continuous derivative.\\

\begin{figure}[t]
\centerline{\includegraphics[height=0.28\textheight]{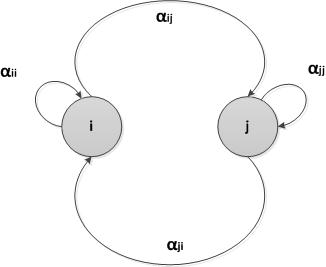}}
\caption{State Transition Probability between two sites $S_i$ and $S_j$}
\label{stateTrans}
\end{figure}

\noindent \textbf{Exchange of Local Function Estimate:}  Each site locally computes the loss based on its attributes and then gossips with the neighbors to get information on other attributes.
On receiving an update from a neighbor, the site re-evaluates $J_i$ by forming a component-wise convex combination of its old vector and the values in the messages received from neighbors i.e.   
 $J_i^{t+1}=\alpha_{ii}(X_i W_i^T) + \alpha_{ji} (X_j W_j^T)$. 
It is interesting to note that $\alpha_{il}, 0 \le \alpha_{il} \le 1$, is a non-negative weight that captures the fraction of information site $i$ is willing to share with site $l$. The choice of $\alpha_{il}$ may be deterministic or randomized and may or may not depend on the time $t$ \cite{Kempe_03}. The $k\times k$ matrix $A$ comprising of $\alpha_{il}, 1 \le i \le k, 1 \le l \le k$ is a ``stochastic" matrix such that it has non-negative entries and each row sums to one. More generally, this reflects the state transition probabilities between sites. Figure~\ref{stateTrans} illustrates the state transition between two sites $S_i$ and $S_j$. 

Another interpretation of the diffusion of $J_i$ amongst the neighbors of $i$ involves drawing analogies from Markov chains -- the diffusion is mathematically identical to the evolution of state occupation probabilities. Furthermore, a simple vector equation can be written for updating $J_i^t$ to $J_i^{t+1}$ i.e. $J_i^{t+1} = A(i) (J_i^t)_{N_i}$ where $A(i)$ corresponds to the row $i$ of the matrix $A$ and $(J_i^t)_{N_i}$ is a matrix that has $|N_i|$ rows (each row corresponding to a neighbor of Site $S_i$) and $n$ columns (each column corresponding to all the instances). More generally, $\mathcal{J}^{t+1} = A \mathcal{J}^{t}$ where $\mathcal{J}^{t+1}$ is a $k \times n$ matrix storing the local function estimates of each of the $n$ instances at site $k$ and $A$ is the $k \times k$ transition probability matrix corresponding to all the sites. It follows that $lim_{t \rightarrow \infty} A^t$ exists and this controls the rate of convergence of the algorithm. 

We introduce the notion of \emph{average function estimate} in the network $\vec{J_i^t} = \sum_i \frac{J_i^t}{k}$ which allocates equal weight to all the local function estimates and serves as a baseline against which individual sites $S_i$'s can compare their performance. Philosophically, this also implies that each local site should at least try to attain as much information as required to converge to the average function estimate. Since $\sum_i{\alpha_{ij}}=1$, this estimate is invariant. 

The $A$ matrix has interesting properties which allow us to show that convergence to $\vec{J_i^t}$ occurs. One such property is the Perron-Frobenius theory of irreducible non-negative matrices. We state the theorem here for continuity.

\begin{thm} \textbf{Perron-Frobenius \cite{Varga_62}}
Let $A$ be a positive, irreducible matrix such that the rows sum to 1. Then the following are true:
\begin{enumerate}
\item The eigenvalues of $A$ of unit magnitude are the $k$-th roots of unity for some $k$ and are all simple.
\item The eigenvalues of $A$ of unit magnitude are the $k$-th roots of unity if and only if $A$ is similar under a permutation to a $k$ cyclic matrix
\item all eigenvalues of $A$ are bounded by 1.
\end{enumerate}
\end{thm}

Since the eigenvalues of $A$ are bounded by 1, it can be shown that $J_i^t$ converges to the average function estimate $\vec{J_i^t}$ if and only if -1 is not an eigen value \cite{Varga_62}. 
Let $\lambda_n \le \lambda_{n-1} \le \cdots \le \lambda_2 < \lambda_1 =1$ be the eigenvalues of $A$ with $\lambda_1 = 1$. Also assume that $\gamma (A) = \text{max}_{i>1} |\lambda_i|$. It can be shown that $\parallel J_i^{t+1} - \vec{J}_i^{t} \parallel^2 \le \gamma^2 \parallel J_i^{t} - \vec{J}_i^{t} \parallel$. If $\gamma=1$, then system fails to converge \cite{Varga_62}, \cite{Cybenko_89}. \\

\noindent \textbf{Local Stochastic Gradient update} is done as follows: $W_i^{t+1} = W_i^t - \eta_i^t s_i^t$ where $s_i^t=\frac{\partial J_i^t}{\partial W_i^t} (X_r, W_i^t), X_r \in \mathbb{R}^{m_i}$ is the estimated gradient, $W_i^t$ is the weight vector and $\eta_i^t$ is the learning rate at node $i$ at time t.

%
%

\subsection{Convergence}
\label{sec:algterm}
The proof of convergence of the algorithm makes use of the following concept: In the distributed setting, the process of information exchange between $k$ sites can be modeled as a non-stationary Markov chain. A non-stationary Markov chain is weakly ergodic if the dependence on the state distribution vanishes as time tends to infinity \cite{Tsitsiklis_86}. A detailed discussion regarding convergence of the algorithm is presented here. \\


\noindent First, we make the following assumptions about the cost function $J$:
\begin{asm}{}
\begin{enumerate}
\item There holds $J(W^t) \ge 0$, for every $W^t \in R^m$
\item \textbf{Lipschitz Continuity of $\nabla J$: } The function $J$ is continuously differentiable and there exists a constant $K_1$ such that 
\begin{equation}
\parallel \nabla J(W^{t_1}) - \nabla J(W^{t_2}) \parallel \le K_1 \parallel W^{t_1} - W^{t_2} \parallel, \forall \text{ } W^{t_1}, W^{t_2} \in \mathbb{R}^m.
\end{equation}
\item If $J$ satisfies the Lipschitz condition above, then
\begin{align*}
\label{desLips}
J(W^{t_1}+W^{t_2}) &\le J(W^{t_1}) + (W^{t_2}) \nabla J(W^{t_1})^{'} + \frac{K}{2} \parallel W^{t_2}\parallel_2^2,  
					&\text{for all }W^{t_1}, W^{t_2} \in \mathbb{R}^m.
\end{align*}
\end{enumerate}
\end{asm}
\noindent In our algorithm, the vector $W^t$ is split over sites $S_1, S_2, \cdots, S_k$. The attributes at site $S_i, (1 \le i \le k) $ are updated according to the following equation:
\begin{equation}
W_i^{t+1} = W_i^t - \eta_i^t s_i^t
\end{equation} 
where $\eta_i^t$ is the step size and $s_i^t$ is the descent direction at site $S_i$. Let $T^i$ be the set of times when processor $i$ makes an update. It is assumed that $s_i^t=0$ when $t \notin T^i$. For times $t \in T^i$, we assume that the update direction is such that the cost function decreases and $s_i^t$ has the opposite sign from $\nabla J_i(W_i^{t})$. The underlying deterministic gossip algorithm is described by:
\begin{equation}
\label{gossip}
W_{i}^{t+1} = \sum_{\{i|t \in T^{i} \}} \alpha_{ii} W_{i}^{t} + \sum_{\{j|t \in T^{i} \}} \alpha_{ij} W_{j}^{t} \end{equation}
\noindent where the coefficients $\alpha$'s  are non-negative scalars.

\begin{ex}
Let $S_i$ and $S_j$ be the only two sites communicating with each other. Then equation~\ref{gossip} reduces to 
\begin{align*}
W_{i}^{t+1} &= \alpha_{ii} W_{i}^{t} + \alpha_{ij} W_{j}^{t} \\
			&= \alpha_{ii} (W_{i}^{t-1} - \eta_i^{t-1} s_i^{t-1}) + \alpha_{ij} (W_{j}^{t-1} - \eta_j^{t-1} s_j^{t-1}) \\
			&= (\alpha_{ii} W_{i}^{t-1} + \alpha_{ij}W_{j}^{t-1}) - ( \alpha_{ii} \eta_i^{t-1} s_i^{t-1} + \alpha_{ij}\eta_j^{t-1} s_j^{t-1}) \\
			&= \cdots \\			
			&= (\alpha_{ii} W_{i}^{1} + \alpha_{ij}W_{j}^{1}) - (\alpha_{ii} (\eta_i^{t-1} s_i^{t-1}+\eta_i^{t-2} s_i^{t-2} \cdots +\eta_i^{1} s_i^{1} ) \\
			 &+ \alpha_{ij} (\eta_j^{t-1} s_j^{t-1} + \eta_j^{t-2} s_j^{t-2} + \cdots + \eta_j^{1} s_j^{1}) )	 
\end{align*} 
\end{ex}

Hence, by induction it can be shown that:
\begin{equation}
\label{gos1}
W_{i}^{t} = \sum_{j=1}^{k} \alpha_{ij}W_{j}^{1} + \sum_{\tau=1}^{t-1}\sum_{j=1}^{k} \alpha_{ij}\eta_j^{\tau} s_j^{\tau}
\end{equation}
It is also assumed that there exist positive constants $K_4$ and $K_5$ such that step-sizes $\eta_i^{t}$ are bounded as follows:
$\frac{K_4}{t} \le \eta_i^{t} \le \frac{K_5}{t}$. Furthermore, the following assumptions hold true:

\begin{asm}
\begin{enumerate}
\item $\forall i, j$ and $0 \le \tau \le t$, $0 \le \alpha_{ij} \le 1$.
\item For any $i, j$ and $\tau \ge 0$, the limit of $\alpha_{ij}$ as $t$ tends to infinity exists and is the same for all $i$ and is denoted by $\alpha_{i}$
\item There exists some $\eta > 0$ such that $\alpha_{j} \ge \eta $ and $\forall j \in \{1, \cdots, k\}$ and $\tau \ge 0$
\item There exists constants $A > 0$ and $\rho \in (0,1) $ such that $|\alpha_{ij} - \alpha_{j}| \le A \rho^{t-\tau}, \forall t> \tau >0$
\end{enumerate}
\end{asm}

\begin{asm}
\label{asm1}

\noindent \textbf{Descent Lemma \cite{Bertsekas_97} at each site:} \\ 
(a) For every $i$ and $t$ we have,
\begin{equation}
s_i^t \nabla J_i(W_i^{t}) \le 0.
\end{equation}

\noindent (b) There exist positive constants $K_2$ and $K_3$ such that
\begin{equation}
K_2 |\nabla J_i(W_i^{t})| \le |s_i^t| \le K_3 |\nabla J_i(W_i^{t})|
\end{equation}
\end{asm}


\noindent Let $\mathcal{S}(t)$ be the set of random variables defined by: 
$\mathcal{S}(t)=\{s_i^{\tau}| i \in \{1, \cdots,k \}, \tau < t \}$. The variables in $\mathcal{S}(t)$ are the only sources of randomness upto the time $t$ at site $i$. The set $\mathcal{S}(t)$ is also a representation of the entire history of the algorithm upto the moment that the update directions $s_i^{\tau}$ are generated.

\begin{asm}
\label{as2}

\noindent \textbf{Stochastic Descent Lemma \cite{Bertsekas_97} at each site:} \\
There exist positive constants $K_6$, $K_7$ and $K_8$ such that: \\
(a) \begin{equation}
\label{descent01}
\nabla J(W_i^{t})' E[s_i^t|\mathcal{S}(t)] \le - K_6 \parallel \nabla J(W_i^{t}) \parallel^2, \forall t \in T^i.
\end{equation} \\
(b)\begin{equation}
\label{descent02}
 E[\parallel s_i^t \parallel^2|\mathcal{S}(t)] \le K_7 \parallel \nabla J(W_i^{t}) \parallel^2 + K_8,  \forall t \in T^i.
\end{equation} 
\end{asm}
\noindent \ref{asm1} implies that the expected direction of the update given the past history is in the descent direction. In \ref{as2} the presence of constant $K_8$ in the inequality allows the algorithm to make non-zero updates even when the minimum has been reached. \\

\begin{asm}

\noindent \textbf{Partial Asynchronism \cite{Bertsekas_97})} \\
There exists a positive integer $B$ such that: \\
(a) For every $i$ and for every $t \ge 0$ at least one of the elements of the set $\{t,t+1,\cdots,t+B-1\}$ belongs to $T^i$. \\
(b) There holds $\text{max}\{0,t-B+1\} \le \tau_{ij}^t \le t$, for all $i$ and $j$ and $t \ge 0$.
\end{asm}
\noindent Finally, for completion we introduce the notions of \emph{martingales} and the martingale convergence theorem(s) which are required for the proofs in this appendix. 

A martingale is a model of a fair game where knowledge of past events never helps predict the mean of the future winnings. In general, a martingale is a stochastic process for which, at a particular time in the realized sequence, the expectation of the next value in the sequence is equal to the present observed value even given knowledge of all prior observed values at a current time\footnote{http://en.wikipedia.org/wiki/Martingale$\_$(probability$\_$theory)}. A formal definition (using measure theory \cite{Tao_11}) is given below:

Let $(\sigma, \mathcal{F}, P)$ be a probability space. A martingale sequence of length $n$, is a sequence $X_1, X_2, \cdots, X_n$ of random variables and corresponding sub-$\sigma$ fields $\mathcal{F}_1, \mathcal{F}_2, \cdots, \mathcal{F}_n$ that satisfy the following relations:
\begin{itemize}
\item Each $X_i$ is an integrable random variable which is measurable with respect to the corresponding $\sigma$-field $\mathcal{F}_i$.
\item The sigma fields are increasing $F_i \subset F_{i+1}$ for every $i$
\item For every  $i \in [1,2,\cdots,n-1]$, we have the relation, $X_i = E[X_{i+1}|\mathcal{F}_i]$ almost everywhere $P$. 
\end{itemize}

Along the same lines,
\begin{itemize}
\item A \emph{submartingale} is defined as: for every  $i$, $X_i \le E[X_{i+1}|\mathcal{F}_i]$ almost everywhere $P$ and 
\item A \emph{supermartingale} is defined as: for every  $i$, $X_i \ge E[X_{i+1}|\mathcal{F}_i]$ almost everywhere $P$. 
\end{itemize}

Martingale convergence theorem is a special type of theorem since the convergence follows from the structural properties of the sequence of random variables. The Supermartingale Convergence theorem and a variant used in proofs is presented next. \\

\noindent \textbf{Supermartingale Convergence Theorem \cite{Bertsekas_97}: } Let $\{Y_i\}$ be a sequence of random variables and let $\{\mathcal{F}_i\}$ be a sequence of finite sets of random variables such that $F_i \subset F_{i+1}$ for each $i$. Suppose that:
\begin{itemize}
\item Each $Y_i$ is non-negative
\item For each $i$, we have $E[Y_i] < \infty$
\item For each $i$, we have $E[Y_{i+1}|\mathcal{F}_i] \le Y_i$ with probability 1. 
\end{itemize}
Then there exists a non-negative random variable $Y$ such that the sequence of $\{Y_i\}$ converges to $Y$ with probability 1.\\

\noindent An extension of the above theorem, can be stated as follows:

Let $\{Y_i\}$ and $\{Z_i\}$ be two sequence of random variables. Let $\{\mathcal{F}_i\}$ be a sequence of finite sets of random variables such that $F_i \subset F_{i+1}$ for each $i$. Suppose that:
\begin{itemize}
\item The random variables $Y_i$ and $Z_i$ are non-negative.
\item There holds $E[Y_{i+1}|\mathcal{F}_i] \le Y_i+Z_i, \forall i$ with probability 1.
\item There holds $\sum_{i=1}^{\infty} E[Z_i] < \infty$.
\end{itemize}
Then there exists a nonnegative random variable $Y$ such that the sequence $\{Y_i\}$ converges to $Y$ with probability 1. \\

\noindent \textbf{Proposition 1.0. Convergence of the DFE Algorithm } Under the Assumptions 1.0-3.0, there exists some $\eta^0 > 0$, such that if $0 < \eta_i^t < \eta^0$, then:
\begin{enumerate}
\item $\lim_{t \to \infty} J(W_i^t)$ exists and is the same for all $i$ with probability 1.
\item $\lim_{t \to \infty} (W_i^t - W_j^t)=0$ with probability 1 and in the mean square sense.
\item For every $i$, $\lim_{t \to \infty}\nabla J(W_i^t) = 0$.
\item Suppose that the set $\{W|J(W) \le C\}$ is bounded for every $C \in \mathcal{R}$; then there exists a unique vector $W^{*}$ at which $J$ is minimized and this is the unique vector at which $\nabla J$ vanishes. Then $W_i^t$ converges to $W^{*}$ for each $i$ with probability 1.
\end{enumerate}

\noindent \textbf{Proof.}
Without loss of generality, assume that $\eta_i^{t} = \frac{1}{t}, \forall i, t$. 

We note that the underlying gossip protocol illustrated by equation~\ref{gossip} has a simple structure but is not easy to manipulate in algorithms primarily because we have one such equation for each $i$ and they are generally coupled. Thus we need to keep track of vectors $W_1^t, W_2^t, \cdots, W_k^t$ simultaneously. Analysis would be simpler if we could associate one single vector $\mathcal{W}^t$ that summarizes the information contained in $W_i^t$'s. Let $\mathcal{W}^t$ be defined as follows:
\begin{equation}
\label{avgGos}
\mathcal{W}^t = \sum_{i=1}^k \alpha_{i} W_i^1 + \sum_{\tau=1}^{t-1} \sum_{i=1}^{k} \alpha_{i} \eta_i^{\tau} s_i^{\tau}
\end{equation}
The interpretation of vector $\mathcal{W}^t$ is quite interesting in the following sense -- if the sites stopped performing updates at time $\bar{t}$, but keep communicating and forming convex combinations of their states using the gossip protocol, they will asymptotically agree and the vector they agree upon is $\mathcal{W}^t$. Finally, $\mathcal{W}^{t+1}=\mathcal{W}^t + \sum_{i=1}^k \alpha_{i} \eta_i^{t} s_i^{t}$.

Define also the following:
\begin{equation}
\label{bee}
b^t = \sum_{i=1}^k \parallel s_i^{t} \parallel, t \ge 1.
\end{equation}

\begin{equation}
G^t = - \sum_{i=1}^k \nabla J(W_i^t)' \alpha_{i} s_i^{t}, \text{ } t \ge 1.
\end{equation}


\noindent \textbf{Lemma 1.0}
(a) If $t \in T^i$, then 
\begin{equation}
E[ G^t | S(t)] \ge K_6 \lambda \sum_{\{i|t \in T^i\}} \parallel \nabla J(W_i^{t})' \parallel^2 \ge 0,
\end{equation}
where $\alpha_{ii}>\lambda, \forall i \in \{1,\cdots,k\} \text{ and }\lambda > 0$. \\ 
(b) If $t \ge 1$, then 
\begin{equation}
\label{boundbt}
E[(b^t)^2 |  S(t)] \le A_1 E[ G^t | S(t)] + A_2 
\end{equation}
where $A_1=\frac{k K_7}{\lambda K_6}$ and $A_2=k^2 K_8$

\noindent \textbf{Proof:} Using Assumption 2.0b, Equation~\ref{descent01} and the fact that $s^i=0, t \notin T^i$, we have:

\begin{align*}
E[ G^t | S(t)] &= - \sum_{i|t \in T^i} \nabla J(W_i^t) \alpha_{i} E [s_i^t | S(t)] \\
			&\ge \sum_{i|t \in T^i} K_6 \parallel \nabla J(W_i^t) \parallel^2 \alpha_{i} \\
			&\ge \lambda K_6 \sum_{i|t \in T^i} \parallel \nabla J(W_i^t) \parallel^2
\end{align*}

\noindent This proves part (a) of the Lemma. Applying Equation~\ref{descent02} we obtain,

\begin{align*}
E[(b^t)^2 |  S(t)] &= E [(\sum_{i=1}^k \parallel s_i^t \parallel)^2|S(t)] \\
				   &\le k \sum_{i=1}^k E[\parallel s_i^t \parallel^2|S(t)] \\
				   &\le k \sum_{i=1}^k (K_7 \parallel \nabla J(W_i^{t}) \parallel^2 + K_8) \\
				   &\le \frac{k K_7}{\lambda K_6} E[ G^t | S(t)] + k^2 K_8
\end{align*}
where the last inequality uses the proof in part (a). Q.E.D.

\noindent \textbf{Lemma 2.0} For every $t \ge 1$, we have 
\begin{equation}
\parallel \mathcal{W}^t - W_i^t \parallel \le A \sum_{\tau=1}^{t-1} \frac{1}{\tau} \rho^{t-\tau}b^{\tau}
\end{equation}
\noindent where $A>0$, $\rho \in (0,1)$ 

\noindent \textbf{Proof: }Subtracting Equation~\ref{avgGos} from Equation~\ref{gos1} we have
\begin{equation}
 \mathcal{W}^t - W_i^t = \sum_{\tau=1}^{t-1} \sum_{j=1}^{k} \frac{1}{\tau} [\alpha_{j} - \alpha_{ij}] s_j^t 
\end{equation}
Furthermore, using Assumption 1.1 (a) and definition~\ref{bee} we obtain,
\begin{align*}
\parallel \mathcal{W}^t - W_i^t \parallel &\le \sum_{\tau=1}^{t-1} \frac{1}{\tau} \sum_{j=1}^{k} A \rho^{t-\tau} \parallel s_j^{\tau} \parallel \\
 & \le  A \sum_{\tau=1}^{t-1} \frac{1}{\tau}  \rho^{t-\tau}  b^{\tau} [Q.E.D]
\end{align*}

Using the fact that $\mathcal{W}^{t+1}=\mathcal{W}^t + \sum_{i=1}^k \alpha_{i} \eta_i^{t} s_i^{t}$ and Assumption 1.0 (3) we obtain:

\begin{align*}
J(\mathcal{W}^{t+1}) &= J(\mathcal{W}^t + \sum_{i=1}^k \alpha_{i} \eta_i^{t} s_i^{t}) \\
					 &\le J(\mathcal{W}^t ) + \sum_{i=1}^k \alpha_{i} \eta_i^{t} s_i^{t} \nabla J(\mathcal{W}^{t+1})^{'} + \frac{K}{2}\parallel \sum_{i=1}^k \alpha_{i} \eta_i^{t} s_i^{t} \parallel^2_2 \\
					 &\le J(\mathcal{W}^t ) + \frac{1}{t} \sum_{i=1}^k \alpha_{i} s_i^{t}  \nabla J(W_{i}^{t+1})^{'} + \frac{1}{t} \sum_{i=1}^k \alpha_{i} s_i^{t} ( \nabla J(\mathcal{W}^{t+1})^{'} -\nabla J(W_{i}^{t+1})^{'})+ \frac{K}{2t^2} \parallel s_i^t \parallel_{2}^{2} \\
					 &\le J(\mathcal{W}^t ) - \frac{1}{t} G(t)  + \frac{1}{t} \sum_{i=1}^k \alpha_{i} s_i^{t} K_1 \parallel \mathcal{W}^{t+1} - W_{i}^{t+1} \parallel + \frac{K}{2t^2} (b^t)^2 \\
					 &\le J(\mathcal{W}^t ) - \frac{1}{t} G(t) + \frac{K_1}{t} \sum_{i=1}^k \alpha_{i} s_i^{t} A \sum_{\tau=1}^{t-1} \frac{1}{\tau}  \rho^{t-\tau}  b^{\tau} + \frac{K}{2t^2} (b^t)^2 \\
					 &\le J(\mathcal{W}^t ) - \frac{1}{t} G(t) + \frac{K_1 A}{t} b^t \sum_{\tau=1}^{t-1} \frac{1}{\tau}  \rho^{t-\tau}  b^{\tau} + \frac{K}{2t^2} (b^t)^2 \\
					 &\le J(\mathcal{W}^t ) - \frac{1}{t} G(t) + K_1 A \sum_{\tau=1}^{t-1} \frac{1}{t \tau}  \rho^{t-\tau}  b^{\tau} b^{t} + \frac{K}{2t^2} (b^t)^2 \\
					&\le  J(\mathcal{W}^t ) - \frac{1}{t} G(t) + K_1 A \sum_{\tau=1}^{t-1} \rho^{t-\tau} (\frac{(b^{\tau})^2}{\tau^2} + \frac{(b^t)^2}{t^2}) + \frac{K}{2t^2} (b^t)^2 \\					&\le J(\mathcal{W}^t ) - \frac{1}{t} G(t) + A_3 \sum_{\tau=1}^{t} \rho^{t-\tau} \frac{(b^{\tau})^2}{\tau^2} \numberthis \label{ineql1} \\
\end{align*}

\noindent where $A_3 = \frac{K_1 A}{1 - \rho} + \frac{K}{2}$ \\

\noindent \textbf{Lemma 3.0} There holds 
\begin{equation}
\sum_{t=1}^{\infty} \frac{1}{t} E[G^t] < \infty
\end{equation}

We take expectations of both sides of the above inequality and use Equation~\ref{boundbt} to bound $E[(b^t)^2]$. This yeilds:
\begin{equation}
\label{eqnSum}
E[J(\mathcal{W}^{t+1})] \le E[J(\mathcal{W}^t)] - \frac{1}{t} E[G^t] + A_3 \sum_{\tau=1}^{t} \rho^{t-\tau} \frac{1}{\tau^2} (A_1 E[G^{\tau} + A_2]).
\end{equation} 

Let $t=1,2,\cdots, \bar{t}$ and add the resulting inequalities from Equation~\ref{eqnSum}. Then,

\begin{align*}
E[J(\mathcal{W}^{\bar{t}+1})] &\le J(\mathcal{W}^1) - \sum_{t=1}^{\bar{t}} \frac{1}{t} E[G^t] + A_2 A_3 \sum_{t=1}^{\bar{t}} \sum_{\tau=1}^{t} \rho^{t-\tau} \frac{1}{\tau^2} + A_1 A_3 \sum_{t=1}^{\bar{t}} \sum_{\tau=1}^{t} \rho^{t-\tau} \frac{1}{\tau^2} E[G^{\tau}]\\
   &= J(\mathcal{W}^1) - \sum_{t=1}^{\bar{t}} \frac{1}{t} E[G^t] (1 - A_1 A_3 \frac{1}{t} \sum_{t=1}^{\bar{t}} \rho^{t-\tau} ) + A_2 A_3 \sum_{t=1}^{\bar{t}} \sum_{\tau=1}^{t} \rho^{t-\tau} \frac{1}{\tau^2}\\
   &\le J(\mathcal{W}^1) - \sum_{t=1}^{\bar{t}} \frac{1}{t} E[G^t] (1 - \frac{A_1 A_3}{t(1-\rho)}) + A_2 A_3 \sum_{\tau=1}^{\bar{t}} \frac{1}{\tau^2(1-\rho)}
\end{align*}
The term $A_2 A_3 \sum_{\tau=1}^{\bar{t}} \frac{1}{\tau^2(1-\rho)}$ is bounded since the infinite sum $\sum_{\tau=1}^{\infty} \frac{1}{\tau^2(1-\rho)}$ is bounded. If $\sum_{t=1}^{\infty} \frac{1}{t} E[G^t]=+\infty$, then the right hand side would equal $-\infty$. However, the left hand side is non-negative. This proves the lemma.  [Q.E.D] \\

\noindent \textbf{Lemma 4.0} The sequence $\{J(\mathcal{W}^t\}$ converges with probability 1. \\

\noindent \textbf{Proof:} Taking conditional expectation of inequality~\ref{ineql1}, conditioned on $\mathcal{S}(t)$ and using Lemma 1.0 we have,

\begin{equation}
\label{eqSome1}
E[J(\mathcal{W}^{t+1})|\mathcal{S}(t)] \le  J(\mathcal{W}^{t}) + A_3 \sum_{\tau=1}^{t} \rho^{t-\tau} \frac{1}{\tau^2} E[(b^{\tau})^2|\mathcal{S}(t)]
\end{equation}

Let $Z(t)=\sum_{\tau=1}^{t} \rho^{t-\tau} \frac{1}{\tau^2} E[{b^{\tau}}^2|\mathcal{S}(t)]$. Using Lemma 1.0 (b) and Lemma 3.0 we have:
\begin{align*}
\sum_{t=1}^{\infty} E[Z(t)] &= \frac{1}{1-\rho} \sum_{t=1}{\infty} \frac{1}{t^2} E[(b^{\tau})^2] \\
							&\le \frac{1}{1-\rho} \sum_{t=1}{\infty} \frac{1}{t^2}(A_1 E[G^{t} + A_2) \\
							&< \infty. \numberthis \label{ineql2} \\
\end{align*}
Using inequalities~\ref{eqSome1} and \ref{ineql2}, a variant of the Supermartingale theorem applies and hence $\{J(\mathcal{W}^t\}$ converges with probability 1.[Q.E.D]

\section{Empirical Evaluation}
\label{sec:expt}

\subsection{Aims}
\label{sec:exptaims}
Our objective is to investigate empirically the utility of the consensus-based
algorithm we have described. 
We use $Model(k,f)$ to denote the model returned the consensus-based
algorithm in Section \ref{sec:alg}
using $k$ nodes in a network, each of which can call on an ILP engine to
construct at most $f$ features. In this section, we compare the
performance of: $Model(N,F)$ $(N > 1)$ with $Model(1,N \times F)$ The latter
effectively represents the model constructed in a non-distributed manner, with
all features present at a single centralised node. For simplicity, we
will call the former the $Distributed$ model and the latter the $Centralised$
model.

We intend to examine if there is empirical support for the conjecture
that the performance of the $Distributed$ model  is better than that of the $Centralised$ model.
We are assuming that the performance of a
model-construction method is given by the pair $(A,T)$
where $A$ is an unbiased estimate of the predictive accuracy of
the classifier, and $T$ is an unbiased estimate of the
time taken to construct a model. In all cases, 
the time taken to construct a model also includes the time taken to identify the
set of features by the ILP engine and the time to compute their values.
When $k > 1$, the time will also include
time for exchanging information. Comparison of
pairs $(A_1,T_1)$ and $(A_2,T_2)$ will simply be lexicographic comparisons.

\subsection{Materials}
\label{sec:exptmat}

\subsubsection{Data}
Data for experiments are in two categories:
\begin{enumerate}
\item \textbf{Synthetic. } We use the ``Trains'' problem
    posed by R. Michalski for controlled experiments. Datasets
    of $1000$ examples
    are obtained for randomly drawn target concepts
    (see ``Methods'' below).\footnote{We note here that we are not concerned
    with large numbers of examples here, since the main investigation is
    concerned with subsets of the feature-space, and not of the data instances.}
    For this we use S.H. Muggleton's random train
    generator\footnote{{\tt http://www.doc.ic.ac.uk/$\mbox{}^\sim$shm/Software/GenerateTrains/}}                                                                                                                                             
    that defines a random process for generating examples. We will use this
    data for controlled experiments to test principal conjecture
    about the comparative performances of $Distributed$ and $Centralised$ models.

\item \textbf{Real. } We report results from experiments conducted using some
    well-studied real world biochemical toxicology problems
    (Mutagenesis \cite{King_96}; Carcinogenesis \cite{King_96a}; and DssTox \cite{something}).
    Our purpose in examining performance
    on the real-data is twofold. First, we intend to see if the use of
    linear models is too restrictive for real problems. Second, we would like to see
    if the results obtained on
    synthetic data are reflected on real-world problems. We note that for
    these problems predictive accuracy is the primary concern.
\end{enumerate}

\subsubsection{Algorithms and Machines}

The DFE algorithm has been implemented on a
Peer-to-Peer simulator, PeerSim \cite{peersim}. 
This software sets up the network by initializing the nodes and the protocols to be used by them.
The newscast protocol, an epidemic content distribution and topology management
protocol is used. Nodes can perform actions on local data as well as communicate with
each other by selecting a neighbour to communicate with
(using an underlying overlay network).
In each communication step, they mutually update their approximations of the
value to be calculated, based on their previous approximations.
The emergent topology from newscast protocol has a very low diameter and is
very close to a random graph (\cite{Jelasity_04},\cite{Jelasity_05}). 

The ILP system used in all experiments is Aleph \cite{Srinivasan_99a}.
The latest version of this program (Aleph 6) is available from the second author.
The Prolog compiler used is Yap (version 6.2.0). The programs are executed on
a dual Quad-Core AMD Opteron 2384 processors equipped with 2.7 GHz processors, 32 GB RAM,
and local storage of $4 \times 146$ GB 15K RPM Serial attached SCSI (SAS) hard disks.

\subsection{Method}
\label{sec:exptmeth}

For the synthetic data, we distinguish between ``simple'' targets
(comprising disjuncts of 1-4 features) and ``complex'' targets (comprising disjuncts of 8--12 features).\footnote{This
distinction between {\em simple\/} and {\em complex\/} is based on results from cognitive psychology which suggest that people find it 
difficult to remember concepts with larger that 7 disjuncts.}
We call this dimension ``Target''. Our method for experiments is straightforward:

\begin{enumerate}
\item For each value of $Target$ \\
      \begin{enumerate}
        \item Randomly draw a target concept from $Target$ \label{meth:drawtarget}
        \item Classify each data instance as $+$ or $-$ using the target concept
        \item Randomly generate a network with $N$ nodes
        \item For each node in the network:
            \begin{enumerate}
                \item Set the number of iterations $T$ and initialize the
                    learning parameter $\eta_i$ for the
                    node. It is assumed that all nodes agree on the initial choice of
                    $T$ and $\eta_i = \eta$. 
               \item Execute the algorithm described in Section \ref{algo} for $T$ iterations
                    and the ILP engine restricted to constructing $F$ features
               \item Record the predictive accuracy $A$ of the (local) model along with
                          the time $T$ taken to construct the model (this includes
                          the feature construction time, and the feature computation time).
                          The pair $(A,T)$ is the performance of the $Distributed$ model for the concept.
            \end{enumerate}
        \item Using a network with a single node:
            \begin{enumerate}
                \item Execute the
                    algorithm described in Section \ref{algo} for $T$ iterations, learning
                    parameter $\eta$, 
                    and the ILP engine restricted to constructing $N \times F$ features
               \item Record the predictive accuracy $A'$ of the mode along with the time
                    taken to construct the model $T'$ (again, this includes the feature construction
                    time and feature computation time).
                    The pair $(A',T')$ is the performance of the $Centralised$ model for the concept.
            \end{enumerate}
      \end{enumerate}
      \item Compare the performances of the $Distributed$ and the Centralised models for the concepts.
\end{enumerate}

\noindent
The following additional details are relevant:

\begin{enumerate}
\item Two sources of sampling variation result with this method. First, variations are possible with
        the target drawn in Step \ref{meth:drawtarget}. Second, to ensure that
        both $Distributed$ and $Centralised$ approaches are constructing features from the same feature-space,
        we employ the
        facility within Aleph of drawing features from an explicitly defined feature space (this
        is specified using a large tabulation of features allowed by the language constraints).
        Although only ``good'' features are retained (see below),
        sampling variations can nevertheless result for both $Distributed$ and $Centralised$ models
        from step of drawing features. In effect, we are performing a randomised search for
        good features within a pre-defined feature space. We report averages for $5$ repetitions
        of draws for the target, and $5$ repetitions of the randomised search for a given target.
\item A target is generated as follows. For simple targets, the number of
    features is chosen randomly from the range 1 to 4. For complex concepts,
    the number of features is randomly chosen from the range 8 to 12. Features are
    then randomly constructed using the ILP engine, and their disjunction constitute the
    target concept.
\item As noted previously, data instances for controlled experiments are drawn
    from the ``Trains" problem. The data generator uses S.H. MuggletonÕs random
    train generator. This implements a random process in which each data instance
    generated contains the complete description of a data object (nominally, a ``train'').
\item An initial set of parameters needs to be set for the ILP engine to describe ``good''
    features.
    These include $C$, the maximum number of literals in any acceptable clause constructed by the
    ILP system; Nodes, the maximum number of nodes explored in any single search conducted by
    the ILP system; Minacc, the minimum accuracy required of any acceptable clause; and Minpos, the
    minimum number of positive examples to be entailed by any acceptable clause.
    $C$ and Nodes are directly concerned with the search space explored
    by the ILP system. Minacc and Minpos
    are concerned with the quality of results returned (they are equivalent to ``precision" and ``support" used
    in the data mining literature). We set $C=4$, Nodes=5000, Minacc=0.75 and
    Minpos=2 for our experiments here. There is no principled reason for these choices, other
    than that they have been shown to work well in the literature.
\item The parameters for the PeerSim simulator include the size of the network,
    degree distribution of the nodes and the protocol to be executed at each node.
    We report here on experiments with a distributed network with $N=10$ nodes.
    Each of these nodes can construct up to $F = 500$ features (per class) and
    the centralised approach can construct up to $N \times F = 5000$ features (per class).
\item The experiments here use the Hinge loss function. The results reported
    are for values of $T$ that the stochastic gradient descent method starts
    to diverge.
\item The learning rate $\eta_i$ remains a difficult parameter in any SGD-based method. There is no
    clear picture on how this should be set. We have adopted the following domain-driven approach.
    In general, lower values of the learning rate imply a longer search. We use three different learning
    rates corresponding to domains requiring high, moderate and low amounts of search (corresponding to
    complex, moderate or simple target concepts). The corresponding learning rates are
    $0.01$, $0.1$ and $1$. We reiterate that there is no prescribed method for deciding these values,
    and better results may be possible with other values.
   The maximum number of iterations $T$ is set to a high value ($1000$). The algorithm
   may terminate earlier, if there are no significant changes to its weight vector.
 \item Since the tasks considered here are binary classification tasks, the performance of the
    ILP system in all experiments will be taken to be the classification accuracy of the
    model produced by the system. By this we mean the usual measure computed from
    a $2 \times 2$ cross-tabulation of actual and predicted classes of instances.
    We would like the final performance measure to be as unbiased as possible by the
    experimental estimates obtained during optimization, and estimates are reported on a holdout set.
\item With results from multiple repetitions (as we have here), it is possible to perform
    a Wilcoxon signed-rank test for both differences and accuracy and differences in time.
    This allows a quantitative assessment of difference in
    performance between the Distributed and the Centralised models. However, results with
    $5$ repetitions are unreliable, and we
    prefer to report on a qualitative assessment, in terms of the average of
    accuracy and time taken. 
\end{enumerate} 

A data instance in each of the real datasets is a molecule, and contains the
complete description of the molecule. This includes: (a) bulk properties, like
molecular weight, logP values etc.; and (b) the atomic structure of the molecule, along
with the bonds between the atoms. For these datasets, clearly
there are no concepts to be drawn, and sampling variation results
solely from the feature-construction process. We therefore only report
on experimental results obtained from repeating the randomised search for
features. Again, estimates of predictive accuracy are obtained from a holdout set.
For mutagenesis and carcinogenesis, each of the 10 computational nodes in the distributed
network construbcts up to  500 features, and the centralised approach constructs up to 5000 features (per class). For
DssTox, we found there were fewer high precision features than the other two datasets.
So the nodes in the distributed network constructs up to 50 features and the centralised node up to 
500 features (per class)

\subsection{Results}
\label{sec:results}

We present first the main results from the from the experiments on synthetic data (shown in Fig.\ref{fig:tab1}).
The primary observations in these experiments are as follows:
(1) On average, as concepts vary, the distributed algorithm appears to achieve higher accuracies
    than the centralised approach, although the differences may not be significant for a randomly
    chosen concept;
(2) On average, as concepts vary, the time taken for model construction by the distributed approach
    can be substantially lower\footnote{Although not apparent in the tabulation, the time is dominated
    by the time for constructing features. As a result, we note that the ratio of times for the
    centralised and distributed approaches need not be (linearly) proportional to the number of nodes
    in the network. For example, the search for a large subset of good features conducted by the
    a centralised approach may take much longer than the search for several small subsets conducted
    by the distributed approach.}; and
(3) The variation in both accuracies and time with the distributed approach due to both changes
    in the concept, or due to repetitions of feature-construction appear to be less than
    the centralised approach.

Taken together, these results suggest that good, stable models can be obtained from the distributed
approach fairly quickly, and that the approach might present an efficient alternative to
a centralised approach in which all features are constructed by a single computational unit.

    

\begin{figure}[htb]
\begin{center}
{\small{
\begin{tabular}{|c|c|c||c|c|}\hline
Model & \multicolumn{2}{|c||}{$Simple$} & \multicolumn{2}{|c|}{$Complex$}\\ \hline
         & Acc.(\%)   & Time (s)        & Acc. (\%)  & Time (s) \\ \cline{2-5}
$Centr.$ & 83.3(14.4) & 451.3(206.0)& 92.0(7.1)  & 222.8(139.6) \\
$Distr.$ & 93.4(10.5) & 54.4(15.3)  & 98.7(1.2)  & 41.6(13.7)\\ \hline
\end{tabular}
}}
\end{center}
\begin{center}
(a) 
\end{center}
\begin{center}
{\small{
\begin{tabular}{|c|c|c||c|c|}\hline
Model & \multicolumn{2}{|c||}{$Simple$} & \multicolumn{2}{|c|}{$Complex$}\\ \hline
         & Acc.(\%)   & Time (s)       & Acc. (\%)  & Time (s) \\ \cline{2-5}
$Centr.$ & 83.6(20.1) & 107.7(50.2)& 95.3(0.6)  & 410.4(8.8) \\
$Distr.$ & 79.9(10.4) & 39.6(9.5)  & 96.6(0.6)  & 52.9(2.7)\\ \hline
\end{tabular}
}}
\end{center}
\begin{center}
(b)
\end{center}

\caption{Results on synthetic data comparing $Centralised$ and $Distributed$ models.
        The results in (a) are averages from repetitions across concepts;
        and in (b) are averages from repetitions of the feature-construction process
        for a randomly drawn concept. In all cases, $Distributed$ denotes
        the model obtained with a $10$ node network, each of which employs
        a randomised search for up to 500 good features. $Centralised$ denotes
        the model obtained with a single node employing a randomised search
        for up to 5000 good features. All randomised searches draw features
        from the same feature-space.}
\label{fig:tab1}
\end{figure}%

What can we expect from the consensus-based learner on the real datasets? Results are in Fig.\ref{fig:tab2}),
and we observe the following:
(1) There is a significant difference in accuracies between the distributed and
    centralised models on two of the datasets ($Canc330$ and $DssTox$). On balance,
    we cannot conclude from this that there is either one of the models is better;
(2) As with the synthetic data, the time for the distributed models is substantially lower.
    As before, the time is dominated by the feature-construction effort. For the real data sets,
    it appears that it is substantially easier to get smaller subsets of good features than larger ones
    (as observed from the substantial difference in the times between the distributed and centralised models); and
(3) Comparisons against the baseline suggest that the use of linear models is not overly restrictive, since the
    models obtained are not substantially worse (predictively speaking) than the ones obtained by ILP in the past.

Again, taken together, these results provide support to the trends observed with the controlled experiments and
suggest that the distributed approach would continue to perform at least as well as the centralised approach
on real data.


%
\begin{figure}[htb]
\begin{center}
{\small{
\begin{tabular}{|c|c|c||c|c||c|c|}\hline
Model & \multicolumn{2}{|c||}{$Mut188$} & \multicolumn{2}{|c||}{$Canc330$} & \multicolumn{2}{|c|}{$DssTox$} \\ \hline
         & Acc.(\%)   & Time (s)       & Acc. (\%)  & Time (s) & Acc. (\%) & Time (s) \\ \cline{2-7}
$Centr.$ & 84.3(2.6) & 141.5(9.1)& 67.6(0.5)  & 553.2(25.1)    & 53.8(0.0) & 377.9(23.3) \\
$Distr.$ & 76.8(0.0) & 2.1(0.2)  & 56.8(0.5)  &  34.9(5.7)     & 61.6(1.0) &  26.9(1.5)   \\[6pt]
Baseline & 84.6(2.6) & --        & 50.4(2.8)  &  --            & 64.7(2.0)  &   --     \\ \hline
\end{tabular}
}}
\end{center}
\caption{Results on real data comparing $Centralised$ and $Distributed$ models.
    The Baseline models are the ones reported in
    \protect{\cite{AshwinJMLR2012}} (these are cross-validation estimates, whereas the
    estimates for $Centralised$ and $Distributed$ models are from holdout sets). No
    estimates of time are reported in that paper.}
\label{fig:tab2}
\end{figure}

\noindent
We turn now to some issues that have been brought out by the experiments:

\begin{description}
    \item[The learning rate $\lambda$.] As will all methods based on stochastic gradient descent,
            the central parameter remains the learning rate $\lambda$. 
                Many strategies have been suggested in literature to automatically adjust the
            learning rates.
        (see for example
        \cite{bottou-2010}, \cite{bottou-bousquet-2011}, \cite{Darken_90}, \cite{Sutton_92}).
        In general, the learning rate on an iteration $\eta_i$ of the algorithm here is of the form $\eta_i = \frac{B}{T^{-\alpha}}$;
        $B=e^{-\lambda T}; 0 < \alpha \le 1$. For experiments reported above, we have used fixed values
        of $\lambda$ based on our assessment of the search required (see the additional details in the Methods section).
        The choice of $\lambda$ is dataset dependent and hence this parameter needs to assigned in some domain-dependent
        manner (as we have done here).
    \item[Convergence Accuracy.] In experiments here (both with synthetic and real data), we have observed that
            the predictive accuracy of the model from the distributed setting is comparable to the predictive accuracy from
            the non-distributed setting. Unlike gains in time which can be expected from a distributed setting,
            it is not evident beforehand
            what can be expected on the accuracy front. This is because the models constructed in the two settings sample different
            sets of features. The results here suggest a conjecture that the consensus-based approach will always converge
            to model that is within some small error bound of the model from a centralised approach with the
            same number of features. We have some reason to believe that this conjecture may hold
            in some circumstances, based on the use of Sanov's theorem \cite{Sanov_57} and related techniques.
\end{description}

\section{Conclusion}
\label{sec:concl}

A particularly effective form of Inductive Logic Programming has been its use
to construct new features that can be used to augment existing descriptors of
a dataset. Experimental studies reported in the literature have repeatedly shown
that the relational features constructed by an ILP engine can substantially assist
in the analysis of data. Models constructed in this way have looked at both classification
and regression, and improvements have resulted in each case. Practical difficulties
have remained to be addressed though. The rich language of first-order logic used
by ILP systems engenders a very large space of possible new features. The resulting
computational difficulties of finding interesting features is not
easily overcome by the usual ILP-based methods of language bias or constraints. In
this paper, we have introduced what appears to be the first attempt at the use
of a distributed algorithm for feature selection in ILP which also has some
provable guarantees of convergence. The experimental results we have presented
suggest that the algorithm is able to identify good models, using significantly lesser
computational resources than that needed by a non-distributed approach.

There are a number of ways in which the work here could be extended further.
Conceptually, we have outlined a conjecture in the previous section
that we believe is worth investigating further. If it is proven to hold, then
this would be a first-of-its-kind result for consensus-based methods. In implementation
terms, we are able to extend the approach we have proposed to other kinds of models
that use convex loss functions, and to consider a consensus-based version of the
SNoW architecture. This latter will give us the ability to partition very large datasets,
and to deal with very large feature-spaces at once. is also not required within the
approach that all computational nodes draw from the same feature space (this was a constraint
imposed here to evaluate the centralised and distributed models in a controlled manner). It may
be both interesting and desirable for nodes to sample from different feature-spaces, or with
different support and precision constraints. Finally,
We need to investigate whether certain kinds of network topologies are better than others (currently,
we impose no control, and use a randomly generated network).
Experimentally, we recognise that
results on more real-world datasets are always desirable: we hope the results here will
provide the impetus to explore distributed feature construction by ILP on many more real datasets. 

{\small{
\section*{Acknowledgements}

H.D. is also an adjunct assistant professor at the Department of Computer Science at IIIT, Delhi and an Affiliated Member of the Institute of Data Sciences, Columbia University, NY. A.S. also holds visiting positions at the School of CSE, University of New South Wales, Sydney; and at the Dept. of Computer Science, Oxford University, Oxford.
}}
%

\bibliographystyle{spmpsci}      
\bibliography{fs}   

\newpage

\end{document}